\documentclass[journal]{IEEEtran}
\usepackage[T1]{fontenc}

\usepackage{cite}

\usepackage{multirow}
\usepackage{graphicx}
\usepackage{amssymb}
\usepackage{amsmath}
\usepackage{amsfonts}
\usepackage{booktabs}
\usepackage{array}
\newcolumntype{L}[1]{>{\raggedright\let\newline\\\arraybackslash\hspace{0pt}}m{#1}}
\newcolumntype{C}[1]{>{\centering\let\newline\\\arraybackslash\hspace{0pt}}m{#1}}
\newcolumntype{R}[1]{>{\raggedleft\let\newline\\\arraybackslash\hspace{0pt}}m{#1}}

\usepackage{lipsum}
\usepackage{changepage}
\usepackage{color,soul}
\usepackage{enumitem}   

\usepackage{pifont}

\usepackage{tikz}

\usepackage{lipsum}

\let\OLDthebibliography\thebibliography
\renewcommand\thebibliography[1]{
  \OLDthebibliography{#1}
  \setlength{\parskip}{0pt}
  \setlength{\itemsep}{0pt plus 0.3ex}
}

\usepackage{array}

\makeatletter
\newcommand{\thickhline}{%
    \noalign {\ifnum 0=`}\fi \hrule height 1pt
    \futurelet \reserved@a \@xhline
}
\newcolumntype{"}{@{\hskip\tabcolsep\vrule width 1pt\hskip\tabcolsep}}
\makeatother

\usepackage{framed,multirow}

\usepackage{latexsym}

\usepackage{url}
\usepackage{xcolor}

\usepackage{hyperref}

\usepackage{bm}

\begin{document}

\title{Critical Assessment of Transfer Learning for Medical Image Segmentation with Fully Convolutional Neural Networks}

\author{Davood Karimi, Simon K. Warfield,~\IEEEmembership{Fellow,~IEEE}, and Ali Gholipour,~\IEEEmembership{Senior Member,~IEEE}

\thanks{Manuscript received May 30, 2020. This study was supported in part by the National Institutes of Health (NIH) grants R01 EB018988, R01 NS106030, R01 NS079788; by a Technological Innovations in Neuroscience Award from the McKnight Foundation; and by the Department of Radiology at Boston Children's Hospital. The content is solely the responsibility of the authors and does not necessarily represent the official views of the NIH or the McKnight Foundation.

D. Karimi, S.K. Warfield, and A. Gholipour are with the Computational Radiology Laboratory of the Department of Radiology at Boston Children's Hospital, and Harvard Medical School, Boston, Massachusetts, USA (email: davood.karimi@childrens.harvard.edu; simon.warfield@childrens.harvard.edu; ali.gholipour@childrens.harvard.edu).}
}

\maketitle


\begin{abstract}

Transfer learning is widely used for training machine learning models. Here, we study the role of transfer learning for training fully convolutional networks (FCNs) for medical image segmentation. Our experiments show that although transfer learning reduces the training time on the target task, the improvement in segmentation accuracy is highly task/data-dependent. Larger improvements in accuracy are observed when the segmentation task is more challenging and the target training data is smaller. We observe that convolutional filters of an FCN change little during training for medical image segmentation, and still look random at convergence. We further show that quite accurate FCNs can be built by freezing the encoder section of the network at random values and only training the decoder section. At least for medical image segmentation, this finding challenges the common belief that the encoder section needs to learn data/task-specific representations. We examine the evolution of FCN representations to gain a better insight into the effects of transfer learning on the training dynamics. Our analysis shows that although FCNs trained via transfer learning learn different representations than FCNs trained with random initialization, the variability among FCNs trained via transfer learning can be as high as that among FCNs trained with random initialization. Moreover, feature reuse is not restricted to the early encoder layers; rather, it can be more significant in deeper layers. These findings offer new insights and suggest alternative ways of training FCNs for medical image segmentation.

\end{abstract}


\IEEEpeerreviewmaketitle

\section{Introduction}

\subsection{Background and motivation}

\IEEEPARstart{D}{eep} learning has made a significant impact in the field of medical image analysis. For semantic segmentation, fully convolutional neural networks (FCNs) have shown to be powerful models \cite{ronneberger2015,milletari2016,kamnitsas2017,karimi2019accurate}. Commonly, FCNs are trained in a supervised manner, i.e., by minimizing some loss function that penalizes the disagreement between the ground truth and predicted segmentations on a set of labeled training images. In medical applications, obtaining ground truth labels is challenging because it requires detailed annotation of large 3D images by domain experts. To address this challenge, a wide range of techniques have been proposed. Some of the main categories of these methods include semi-supervised learning, transfer learning, learning from noisy labels, and learning from computer-generated labels. Recent reviews of these methods for medical image analysis can be found in \cite{cheplygina2019,tajbakhsh2019,karimi2019noisylabel,karimi2020learning}. The focus of this study is on transfer learning.

Broadly speaking, transfer learning refers to any learning strategy that uses the knowledge gained in solving one problem, Problem S, in subsequently solving a separate problem, Problem T. To distinguish transfer learning from related learning strategies such as multi-task learning, it is assumed that Problem T is addressed separately after Problem S \cite{raina2006,pan2009b}. A formal definition, as put forward in \cite{pan2009b}, involves the concepts of \emph{domain} and \emph{task}. A domain $\mathcal{D}$ is defined by a feature space $\mathcal{X}$ and a probability distribution $P(X)$ defined over $\mathcal{X}$. A task $\mathcal{T}$, on the other hand, is defined by a label space $\mathcal{Y}$ and a prediction function $f(x)= P(y | x)$ for $x \in X$ and $y \in Y$. Now, consider a source domain and task ($\mathcal{D}_S$, $\mathcal{T}_S$) and a target domain and task ($\mathcal{D}_T$, $\mathcal{T}_T$), where either $\mathcal{D}_S \neq \mathcal{D}_T$ and/or $\mathcal{T}_S \neq \mathcal{T}_T$.  Transfer learning aims to utilize the knowledge gained in learning $f_S$ in subsequently learning $f_T$. Whereas the above is a formal description of transfer learning, its implementation in practical applications can take many different forms, depending on what kind of information is transferred and how it is utilized in learning $f_T$ \cite{gammerman2013,pan2009b}.

Transfer learning, in its various manifestations, has been widely employed in training deep learning models. Some of the notable examples include studies that aim at learning deep representations that can be re-purposed for other tasks \cite{yosinski2014,donahue2014}, Deep Adaptation Networks for domain adaptation \cite{long2015b}, and few/zero-shot learning \cite{zhang2015,xian2018}. However, for vision applications, the most widely used approach is to pre-train a model on a source domain/task and then fine-tune that same model on the target domain/task \cite{goodfellow2016}. In this approach, the knowledge that is transferred from the source to the target problem is in the form of the values of the network parameters.

Transfer learning has also been used in training deep learning models for various medical image analysis applications \cite{cheplygina2019,tajbakhsh2019}. However, for segmentation, which is the focus of this work, most of the previous studies have been limited to one particular application or dataset, limiting the generalizability of their findings. Furthermore, they have only reported segmentation accuracy measures, without investigating how transfer learning takes place and in what ways the models learned with and without transfer learning differ. This paper aims at filling this gap by presenting a more comprehensive and more in-depth assessment of the effect of transfer learning for FCN-based medical image segmentation.

\subsection{Related works}

Many studies in recent years have used transfer learning for medical image segmentation. Recent reviews of these studies can be found in \cite{cheplygina2019,tajbakhsh2019}. Given our space limitation, here we briefly review some typical examples. In fact, the way the segmentation problem is formulated and the approach used for transfer learning vary greatly between these studies, making some of them less relevant to our work. As an example, one study used transfer learning for segmentation of carotid intima-media boundary and found that transfer learning with a model pre-trained on natural images was useful \cite{tajbakhsh2016}. However, they formulated the segmentation problem as a pixel-wise \emph{classification} task and used a (non-FCN) classification network architecture. Such studies are not directly relevant to our work, which focuses on FCN segmentation models.

One study found that a model trained for liver and kidney segmentation on a dataset of 35 MR images performed very poorly when applied on a second dataset of 45 images, even though the main difference between the two datasets were image size and resolution \cite{valindria2018}. The authors found that fine-tuning the model trained on the first dataset for the segmentation on the second dataset performed equally with training a model from scratch. They proposed using Reverse Classification Accuracy, \cite{valindria2017reverse}, to select the most useful images for annotation in the target domain and showed that with this strategy, using as few as five images in the target domain was sufficient to match the accuracy obtained with all 45 images, both with fine-tuning and with training from scratch.

For brain white matter hyperintensity segmentation in MRI, one study evaluated the effect of transfer learning when source and target domains differed in terms of acquisition protocol \cite{ghafoorian2017}. Compared with training from scratch, transfer learning achieved better results. As the number of training images in the target domain decreased, achieving good performance with transfer learning required limiting the fine-tuning to the top two layers. Similar observations were reported for multiple sclerosis lesion segmentation in multi-site datasets in \cite{valverde2019one}.

A transfer learning method for cross-modality domain adaptation was proposed in \cite{dou2018unsupervised} and successfully applied for segmentation of cardiac CT images using models pre-trained on MR images. The method included a domain adaptation module, based on adversarial training, to map the target data to the source data in feature space. A GAN-based method for mapping the target images to the appearance of source images was proposed in \cite{chen2018semantic}. This method also showed promising results on the segmentation of cross-site chest X-ray datasets.

Even though the studies reviewed above present useful knowledge regarding the effectiveness of transfer learning for medical image segmentation, they are all limited to a single dataset or application. Furthermore, they all lack any analysis of the role of transfer learning beyond the gross segmentation accuracy values. One recent paper reported an in-depth study of transfer learning for medical image analysis \cite{raghu2019transfusion}. However, that study was limited to 2D images and examined transfer learning with models pre-trained on natural images, which is not relevant for 3D medical images. Moreover, that study was dedicated to classification tasks, whereas the work presented in this paper focuses on voxel-wise semantic segmentation.

\subsection{Contributions of this work}

The main contributions of this work include:

\begin{itemize}

\item We experimentally assess the effect of transfer learning in training FCNs for medical image segmentation on a large number of datasets. The difference between source and target domains in our experiments spans a wide range of important factors such as image modality, organ of interest, image quality, and subject age. Overall, our experiments show that although transfer learning reduces the training time on the target task, the improvement in segmentation accuracy is highly task/data-dependent.

\item In trying to explain some of our experimental observations, we show that the representations extracted by the encoder section of the model do not change significantly from their randomly-initialized or pre-trained values during training/fine-tuning. We show that the filters of the encoder section of trained models look random. Using prior studies on neural networks, we explain this behavior by arguing that the responses of such random filters are similar to useful operations such as edge detectors.

\item We further show that it is possible to freeze the filters of the encoder section of the model at their initial random state and train only the decoder section. We show that this training strategy leads to very small or no loss of test accuracy, and may speed up the convergence too.

\item We analyze FCN representations to shed light on the effects of transfer learning on these models. Our analysis shows that there is substantial variability among the final converged models in terms of learned representations throughout the network. In this regard, models trained with transfer learning can be as diverse as models trained from scratch. Moreover, we show that feature reuse is not restricted to the early layers; rather, it can be even more significant in deeper layers, suggesting alternative approaches to model fine-tuning on the target task.

\end{itemize}

\section{Materials and Methods}

\subsection{Data}

Table \ref{table:data} summarizes the information about the datasets used in this work. For all experiments with any of these datasets, we used 70\% of the images for training and validation and the remaining 30\% for test. Our data pre-processing included: 1) resampling images and segmentations for each dataset to an isotropic voxel size; depending on the original voxel spacing of the images in a dataset, the re-sampled voxel size ranged from 0.8 mm to 2.0 mm, 2) intensity normalization: Computed Tomography (CT) images were normalized by linearly mapping the Hounsfield Unit values in the range $[-1000,1000]$ to intensity range $[0,1]$, whereas Magnetic Resonance (MR) images were normalized by dividing each image by the standard deviation of its voxel intensities.

\begin{table*}[!htb]
\footnotesize
 \caption{\small{Summary of the information on the datasets used in this study. The first column shows the names that we use to refer to each dataset throughout the paper. CP stands for brain cortical plate.}}
  \label{table:data}
   \begin{center}
    \begin{tabular}{ L{2.6cm}  C{2.0cm} C{4.8cm} C{1.0cm} C{5.5cm}}
\hline
name & modality & organ & data size & source   \\ \hline
CP- younger fetus & T2 MRI & brain cortical plate & 27 &  In-house (Boston Children's Hospital) \\
CP- older fetus & T2 MRI & brain cortical plate &  15 &   In-house (Boston Children's Hospital) \\
CP- newborn & T2 MRI & brain cortical plate & 558 & \cite{bastiani2019} \\
KiTS & CT & kidney & 300 & \cite{heller2019kits19} \\
LiTS & CT & liver & 130 & \cite{bilic2019liver} \\
Liver-CT & CT & liver & 19 & \cite{heimann2009b} \\
Spleen & CT & spleen & 41 & \url{https://decathlon-10.grand-challenge.org/} \\
Pancreas & CT & pancreas & 281 & \url{https://decathlon-10.grand-challenge.org/} \\
Prostate & MRI & prostate & 32 & \url{https://decathlon-10.grand-challenge.org/} \\
Hippocampus & MRI & hippocampus & 260 & \url{https://decathlon-10.grand-challenge.org/} \\
BRATS & MRI & brain tumor & 484 & \url{https://decathlon-10.grand-challenge.org/} \\
TSC & MRI & Tuberous sclerosis complex lesions & 165 & In-house (Boston Children's Hospital) \\
Liver-MRI-SPIR  & MRI & liver & 20 & \cite{kavur2019} \\
Liver-MRI-DUAL-in  & MRI & liver & 20 & \cite{kavur2019} \\
Liver-MRI-DUAL-out  & MRI & liver & 20 & \cite{kavur2019} \\
\hline
\end{tabular}
  \end{center}
\end{table*}

\subsection{Network architecture and training details}
\label{network}

\begin{figure}[!htb]
  \centering
  \centerline{\includegraphics[width=0.5\textwidth]{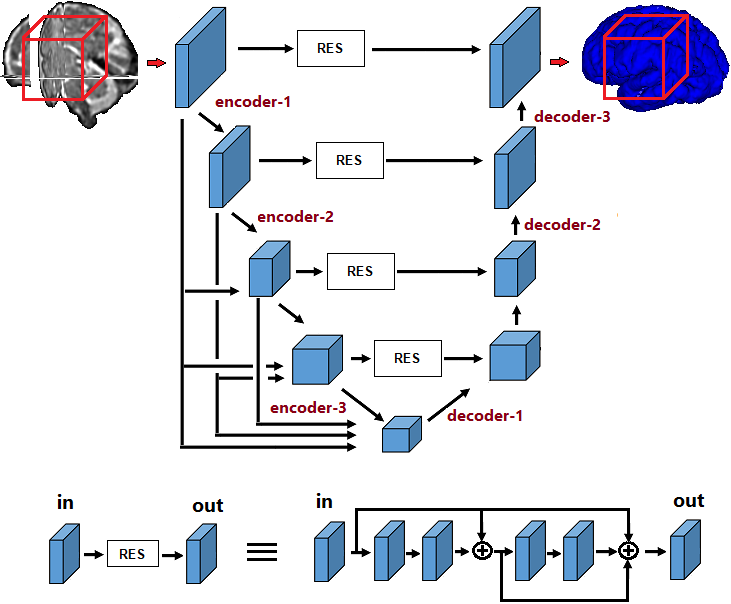}}
\caption{Schematic representation of the our FCN architecture. The lower part of the figure shows details of the residual block.}
\label{fig:network}
\end{figure}

Figure \ref{fig:network} shows the main network architecture used in this study. The overall architecture is similar to the 3D U-Net and V-Net \cite{cciccek2016,milletari2016}, with additional connections between different feature maps in the encoder section of the network. However, in a few of the experiments (indicated explicitly below), we also used the exact original V-Net architecture as explained in \cite{milletari2016}. In Figure \ref{fig:network}, we have marked three convolutional layers in the encoder section and three layers in the decoder section of the network. These are six layers that we will focus on below when we investigate the training dynamics and the effects of transfer learning.

The number of feature maps in the first stage of the network was set to 14, the largest possible on the memory on our GPU. The model accepts $96^3$-voxel image blocks as input, which are sampled from random image locations during training. At test time, a sliding window approach with a 24-voxel overlap between adjacent blocks was used to process an image. In addition to random shifts, other data augmentations used during training included flip, rotation by integer multiples of $\pi/2$, and addition of random Gaussian noise to voxel intensity values. When training from scratch, we use the initialization method proposed in \cite{he2015}. This method initializes convolutional filters with zero-mean Gaussian random variables with a standard deviation of $\sqrt{2/n}$, where $n$ is the number of connections to the convolutional filter from the previous layer. The network was trained by minimizing the negative of the Dice Similarity Coefficient (DSC) between the predicted and target segmentation maps using Adam \cite{kingma2014}. We used an initial learning rate of $10^{-4}$, which was reduced by 0.90 after every 2000 training iterations if the loss did not decrease. 

For our transfer learning experiments, we reduced the initial learning rate by half and fine-tuned all model layers. This has been referred to as deep fine-tuning \cite{tajbakhsh2016}. Some studies, e.g., \cite{ghafoorian2017}, have reported that fine-tuning only certain network layers could be preferable in some applications. However, in our experiments we found that fine-tuning the entire network invariably led to models that were better than or as good as models fine-tuned partially. We use the terms ``training with random initialization" and ``training from scratch", interchangeably, to refer to model training without transfer learning.

To quantify segmentation performance, we mainly use DSC, Average Symmetric Surface Distance (ASSD), and the 95 percentile of the Hausdorff Distance (HD95). In experiments with brain lesion segmentation, we also report the lesion-count F1-score for completeness.

\subsection{Analysis of model training}
\label{cca_section}

Due to their deep hierarchical structure and large number of parameters, deep learning models are considered to be more difficult to interpret and understand than many other machine learning models. Nonetheless, there exist methods for probing the inner workings of these models. In this study, we use some of the recent methods that have been developed for investigating how neural network representations evolve over time and for comparing the representations learned by different networks \cite{raghu2017svcca,morcos2018insights,raghu2019transfusion}.

These methods are based on canonical correlation analysis (CCA) \cite{bach2005}. Given two vectors of random variables, $x \in {\rm I\!R}^n$ and $y \in {\rm I\!R}^m$, CCA seeks projection vectors $u_1 \in {\rm I\!R}^n$ and $v_1 \in {\rm I\!R}^m$ such that the correlation between the projected random variables, $\rho_1= \text{corr}( u_1^Tx , v_1^Ty)$, is maximized. This process can be carried out $\min (m,n)$ times, with the condition that the next pair of projection vectors, $u_i$ and $v_i$, are pairwise-orthogonal to the previously-computed ones. It has been shown that CCA can be used to compare the representations learned by different neural networks \cite{raghu2017svcca,morcos2018insights}. In this setting, elements of vectors $x$ and $y$ correspond to individual neurons of a fully-connected layer or (as in this work) different channels of a convolutional feature map. These random vectors can be sampled by passing data through the network and recording the neuron activation values. In our experiments, we sampled blocks from random locations in the test images and recorded the values of the convolutional layer neurons.

The output of CCA is a set of projection directions, $\{ u_i,v_i \}$, and a measure of how strongly the two representations are correlated along those directions, $\{ \rho_i \}$. In \cite{raghu2017svcca}, it was suggested to use the average of $\rho_i$s as a measure of similarity of two convolutional layers. It was later shown in \cite{morcos2018insights} that the computed directions can vary greatly in terms of the amount of variability in the original data that they explain. Therefore, a weighted average of $\rho_i$s was proposed for estimating the similarity between two convolutional layers: 

\begin{equation} \label{eq:RSIM}
\text{RSIM}(L_1,L_2)= \sum \alpha_i \rho_i,
\end{equation}

\noindent where $L_1$ and $L_2$ denote the convolutional layers being compared and $\alpha_i$ are the normalized weights that are proportional to the amount of variability explained by each direction \cite{morcos2018insights}. In this study, we use this method to compute the similarity of convolutional representations within and across networks to understand the effect of transfer learning.

\section{Results and Discussion}

In this section, we first present the results of a series of experiments to assess the impact of transfer learning in medical image segmentation with FCNs. Each experiment displays a distinct difference between source and target domains.

\subsection{Transfer across imaging modalities}

\begin{figure}[!htb]
  \centering
  \centerline{\includegraphics[width=8.5cm]{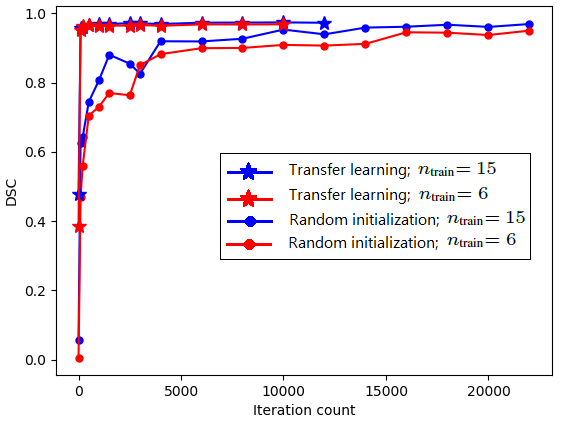}}
\caption{Test DSC as a function of training iteration count for segmentation of the Liver-CT dataset with models trained from scratch and transfer learning.}
\label{fig:tl_silver}
\end{figure}

Here, the organ of interest in the source and target domains is the same, but the imaging modalities are different. The example considered here is liver segmentation. We train a model for segmentation of liver in the pool of the three liver MRI datasets (Table \ref{table:data}). We then fine-tune that model for segmentation of the Liver-CT dataset. The comparison with training from scratch is shown in Table \ref{table:tl_liverct_table} for two sets of experiments with 15 and 6 target training images. In this table, and henceforth in the paper, we use T.L. and R.I. as short for transfer learning and random initialization (i.e., training from scratch), respectively. Figure \ref{fig:tl_silver} shows the test DSC as a function of training iteration count. We see that transfer learning improves the convergence speed. However, in terms of segmentation accuracy, the difference between models trained with transfer learning and learned from scratch is marginal.

\begin{table}[!htb]
\scriptsize
 \caption{\small{Test segmentation accuracy on the Liver-CT dataset for models learned from scratch with random initialization (R.I.) and for transfer learning (T.L.) via fine-tuning a model pre-trained on liver MRI datasets.}}
    \label{table:tl_liverct_table}
\begin{tabular}{ L{1.4cm}  L{0.7cm} C{1.5cm} C{1.5cm} C{1.5cm} }
\thickhline
 &  & DSC & HD95 (mm) & ASSD (mm)   \\ \thickhline
\multirow{2}{*}{\parbox{1.2cm}{$n_{\text{train}}=15$}} & R.I.  & $0.97 \pm 0.01$ & $5.07 \pm 1.94$ & $1.47 \pm 0.33$   \\
& T.L.  & $0.97 \pm 0.01$ & $4.75 \pm 1.81$ & $1.43 \pm 0.33$   \\
\thickhline
\multirow{2}{*}{\parbox{1.2cm}{$n_{\text{train}}=6$}} & R.I.  & $0.95 \pm 0.01$ & $5.47 \pm 2.00$ & $1.61 \pm 0.36$  \\
& T.L.  & $0.96 \pm 0.01$ & $5.25 \pm 2.09$ & $1.56 \pm 0.34$   \\
\thickhline
\end{tabular}
\end{table}

\subsection{Transfer across subject age}

\begin{figure}[!htb]
  \centering
  \centerline{\includegraphics[width=8.5cm]{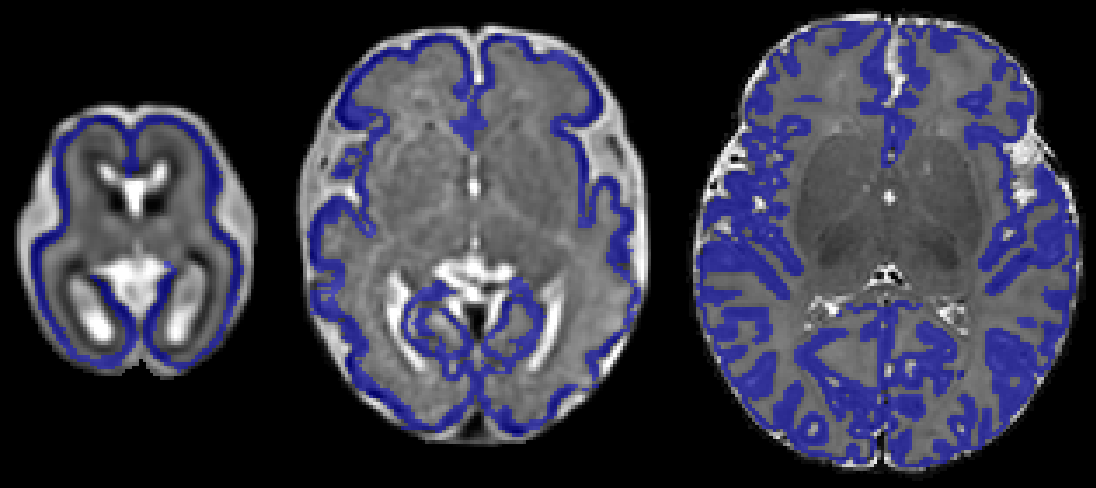}}
\caption{From left to right, example images and segmentations (in blue) from CP- younger fetus, CP- older fetus, and CP- newborn datasets.}
\label{fig:cp_samples}
\end{figure}

Often the source and target domains can be different due to a shift in factors such as subject age or body size. An example of such a shift is represented by the three cortical plate segmentation datasets used in this work. As shown in Figure \ref{fig:cp_samples}, the shape of the cortical plate undergoes significant changes during early brain development.

\begin{figure}[!htb]
  \centering
  \centerline{\includegraphics[width=8.5cm]{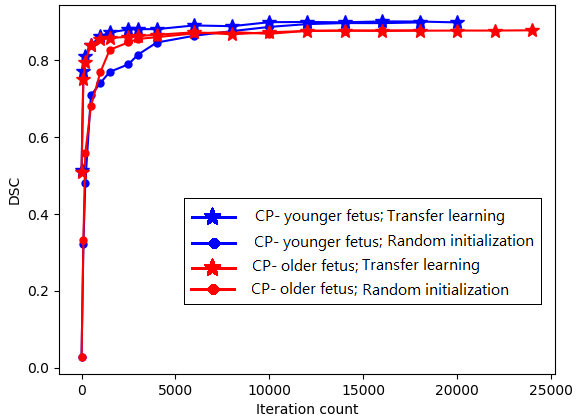}}
\caption{Test DSC as a function of iteration count for segmentation of CP-younger fetus and CP-older fetus datasets with models trained from scratch and transfer learning.}
\label{fig:tl_cp}
\end{figure}

The numbers of images in CP-younger fetus, CP-older fetus, and CP-newborn datasets were, respectively, 27, 15, and 558. We trained a model on CP-newborn, achieving a DSC of 0.93. We then fine-tuned this model on CP-younger fetus and CP-older fetus datasets, both with their entire training set as well as with subsets of 5 images from each dataset. The results are presented in Table \ref{table:tl_cp_table} and Figure \ref{fig:tl_cp}. They show faster convergence with transfer learning. However, improvements in model performance are generally small. The largest improvements were observed for segmentation of CP-older fetus when 5 images were used from the target domain. This suggests that transfer learning may be more effective when source and target domains are more similar and the target training data is smaller.

\begin{table}[!htb]
\scriptsize
 \caption{\small{Test accuracy on CP-younger fetus and CP-older fetus datasets for models learned with R.I. and with T.L. via fine-tuning a model pre-trained on CP-newborn.}}
    \label{table:tl_cp_table}
\begin{tabular}{ L{1.8cm}  L{1.1cm} C{0.5cm} C{1.5cm} C{1.5cm} }
\thickhline
 &  & & DSC & HD95 (mm)   \\ \thickhline
\multirow{4}{*}{\parbox{1.8cm}{CP- younger fetus}} & \multirow{2}{*}{{n train=16}} & R.I.  &  $0.90 \pm 0.03$ & $0.80 \pm 0.02$   \\ 
&  & T.L.  & $0.90 \pm 0.03$ & $0.80 \pm 0.01$   \\ \cline{2-5}
& \multirow{2}{*}{{n train=5}} & R.I.  & $0.88 \pm 0.03$ & $0.86 \pm 0.12$   \\
&  & T.L. & $0.89 \pm 0.03$ & $0.83 \pm 0.06$   \\
 \thickhline
\multirow{4}{*}{\parbox{1.8cm}{CP- older fetus}} & \multirow{2}{*}{{n train=10}} & R.I.  &  $0.82 \pm 0.05$ & $1.02 \pm 0.19$   \\
& & T.L.  & $0.82 \pm 0.05$ & $0.96 \pm 0.19$   \\
 \cline{2-5}
& \multirow{2}{*}{{n train=5}} & R.I.  & $0.79 \pm 0.05$ & $1.20 \pm 0.26$   \\
& & T.L.  &  $0.82 \pm 0.04$ & $0.97 \pm 0.20$   \\
\thickhline
\end{tabular}
\end{table}

\subsection{Transfer across tasks}

A common scenario arises when a source dataset from the same modality is available but the organ of interest is different between source and target domains. We present two sets of experiments representing this scenario.

The first set of experiments was on segmentation of Pancreas-CT dataset. We trained models from scratch using 150 and 15 training images. We also pre-trained a model on the pool of the other four CT datasets from Table \ref{table:data} and fine-tuned it on the same number of target (Pancreas-CT) images. Comparison of test accuracy for this experiment is shown in Table \ref{table:tl_pancreasct_table}. In this set of experiments, similar to the above, transfer learning improved the convergence speed (plots not shown due to space limitation). The improvement in segmentation accuracy was small when 150 training images were used in the target domain. With only 15 training images, the improvement was more significant, reducing HD95 and ASSD by more than 10\%.

\begin{table}[!htb]
\scriptsize
 \caption{\small{Test segmentation accuracy on the Pancreas- CT dataset for models learned with R.I. and with T.L. via fine-tuning a model pre-trained on the the other four CT datasets from Table \ref{table:data}.}}
    \label{table:tl_pancreasct_table}
\begin{tabular}{ L{1.5cm}  L{0.7cm} C{1.5cm} C{1.5cm} C{1.5cm} }
\thickhline
 &  & DSC & HD95 (mm) & ASSD (mm)   \\ \thickhline
\multirow{2}{*}{\parbox{1.4cm}{$n_{\text{train}}=150$}} & R.I.  & $0.80 \pm 0.07$ & $7.68 \pm 4.45$ & $2.04 \pm 0.50$   \\
& T.L.  & $0.81 \pm 0.07$ & $7.55 \pm 4.24$ & $2.01 \pm 0.43$   \\
\thickhline
\multirow{2}{*}{\parbox{1.2cm}{$n_{\text{train}}=15$}} & R.I.  & $0.70 \pm 0.13$ & $9.12 \pm 2.63$ & $2.55 \pm 0.60$   \\
& T.L.  & $0.74 \pm 0.10$ & $8.11 \pm 2.21$ & $2.23 \pm 0.54$  \\
\thickhline
\end{tabular}
\end{table}

The second set of experiments was on segmentation of brain lesions in the TSC dataset. The TSC dataset included 165 scans from five different centers, with 18-47 scans per center. All scans had been manually annotated in detail. Nonetheless, two or three scans from each center (for a total of 12 scans) were selected for more accurate and detailed annotation by two annotators; these scans were used as test data and the remaining 16-44 scans per center were used for training. We ran two experiments on data from each center: 1) training from scratch, 2) transfer learning by fine-tuning a model pre-trained on the BRATS dataset. We compared these two approaches using all training data from each center as well as using 3 scans from each center. Table \ref{table:tl_tsc_table} shows the accuracy on the 12 test scans. Results show only a marginal improvement when 16-44 training scans were available in the target domain. With only 3 training scans in the target domain, the improvements gained with transfer learning were much larger.

\begin{table}[!htb]
\scriptsize
 \caption{\small{Comparison of test segmentation accuracy on the TSC dataset for models trained with R.I. and with T.L. via fine-tuning a model pre-trained on the BRATS dataset.}}
    \label{table:tl_tsc_table}
\begin{tabular}{ L{2.0cm}  L{1.0cm} C{2.0cm} C{2.0cm} }
\thickhline
 &  & DSC & F1   \\ \thickhline
\multirow{2}{*}{\parbox{1.7cm}{$n_{\text{train}}=16-44$}} & R.I.  & $0.63 \pm 0.14$ & $0.67 \pm 0.18$   \\
& T.L.  & $0.64 \pm 0.14$ & $0.69 \pm 0.18$    \\
\thickhline
\multirow{2}{*}{\parbox{1.2cm}{$n_{\text{train}}=3$}} & R.I.  & $0.48 \pm 0.20$ & $0.50 \pm 0.17$   \\
& T.L.  & $0.60 \pm 0.15$ & $0.64 \pm 0.16$  \\
\thickhline
\end{tabular}
\end{table}

\subsection{Transfer across acquisition protocols}

\begin{figure}[!htb]
  \centering
  \centerline{\includegraphics[width=8.5cm]{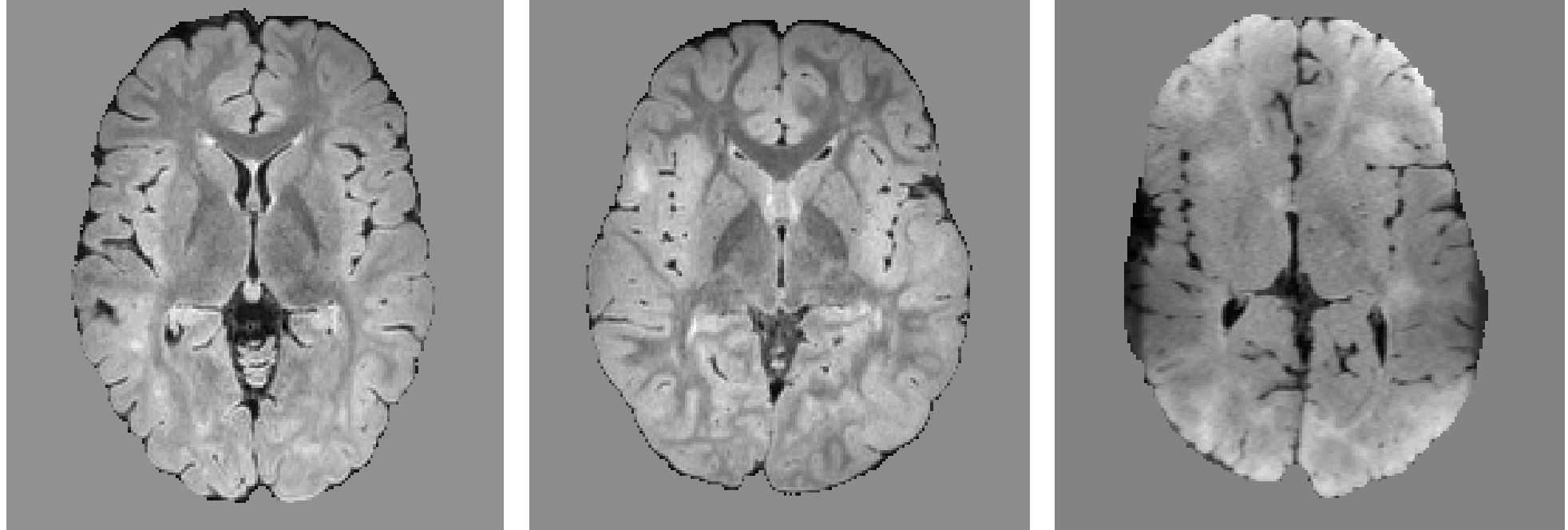}}
\caption{Example FLAIR images in the TSC data from Center 1 (left), Center 3 (middle) and Center 5 (right). Compared to other centers, the FLAIR images from Center 5 had lower tissue contrast and lower effective spatial resolution.}
\label{fig:poor_flair}
\end{figure}

Another recurring theme in medical image data involves varying image quality, such as when different scanners or acquisition protocols are used. As an example of this scenario, we can again refer to the TSC dataset. As we mentioned above, this dataset contained scans acquired at five different centers. For one of these centers, which we refer to as Center 5, the images had reduced gray matter - white matter contrast and resolution compared to the other centers due to the capabilities of the acquisition device. Sample FLAIR images from centers 1, 3, and 5 are shown in Figure \ref{fig:poor_flair}. When we trained and tested models separately on data from each center, the segmentation accuracy on Center 5 was much lower than the other four centers, confirming that the reduced image contrast and resolution resulted in poor model accuracy.

In order to investigate the potential of transfer learning to improve the segmentation accuracy on data from Center 5, we trained models using three strategies: 1) training from scratch on data from Center 5, 2) fine-tuning models pre-trained on each of the other four centers, 3) fine-tuning a model pre-trained on the pool of all data from the other four centers. Table \ref{table:tl_center5_table} shows the results of these experiments, where for the second training strategy we show the average of four trials. The results show a remarkable improvement in the segmentation accuracy of this challenging dataset with transfer learning. Fine-tuning the model trained on the pool of data from all four centers led to better results than fine-tuning the model trained on data from one center.

\begin{table}[!htb]
\scriptsize
 \caption{\small{Test segmentation accuracy on data form Center 5 in the TSC dataset for models trained with R.I. and with T.L. via fine-tuning models pre-trained on data from other centers.}}
    \label{table:tl_center5_table}
\begin{tabular}{ L{3.5cm}   C{2.0cm} C{2.0cm} }
\thickhline
  & DSC & F1   \\ \thickhline
R.I.                          & $0.421$ & $0.393$    \\
T.L. (train on one center)    & $0.577$ & $0.514$  \\
T.L. (train on four centers)  & $\bm{0.582}$ & $\bm{0.525}$  \\
\thickhline
\end{tabular}
\end{table}

\subsection{Investigation of the dynamics of learned representations}

We used the tools described in Section \ref{cca_section} to investigate the dynamics of learned representations to gain a deeper understanding of the effects of transfer learning.

\begin{figure*}[!htb]
  \centering
  \centerline{\includegraphics[width=\textwidth]{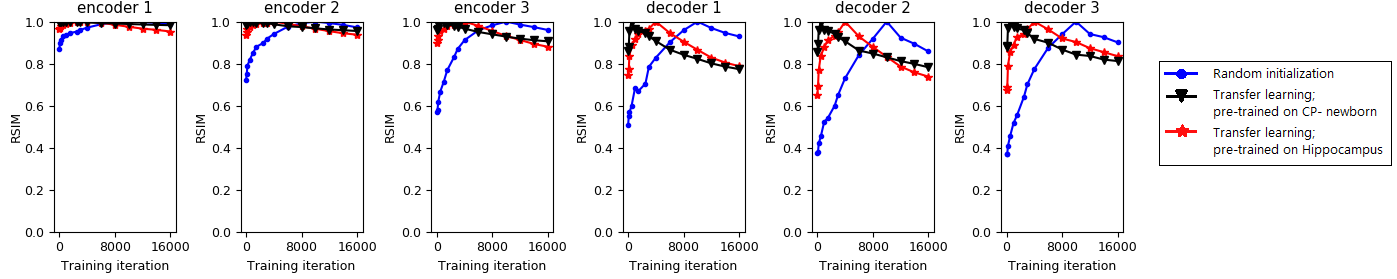}}
\caption{Evolution of learned representations with training for segmentation of CP-younger fetus dataset. Plots show values of RSIM between the representations at each time point with the convergence epoch. Convergence epoch can be identified as the point where RSIM=1.}
\label{fig:dynamics_cp}
\end{figure*}

Figure \ref{fig:dynamics_cp} shows the evolution of learned representations for segmentation of CP-younger fetus dataset for two transfer learning trials as well as training from random initialization. For each experiment, we first chose the ``convergence epoch", which we defined as the epoch when the DSC on the validation set reached within 0.5\% of its maximum. We then computed the similarity of the learned representations between each training epoch and the convergence epoch using Eq. \eqref{eq:RSIM}. Given the large number of convolutional layers in our network, we present this evaluation for the six layers shown in Figure \ref{fig:network}.

Several interesting observations can be made from Figure \ref{fig:dynamics_cp}. First, as expected, the model converged much faster with transfer learning, compared with training from scratch. The convergence epoch can be identified as the point where RSIM=1 since we are comparing the representations in each epoch with those at the convergence epoch. The second observation is that the representations in all layers continue to change significantly well after the model has converged. It is worth noting that after the convergence epoch, the training and test accuracies changed very little. This indicates that the model weight values that can result in a specific test accuracy are far from being unique. This is not an unexpected observation given the very large number of model parameters. The third observation is the effect of the dataset used in pre-training. In this experiment we used two different pre-trained models for transfer learning: one trained on CP-newborn dataset and the other trained on Hippocampus dataset. Figure \ref{fig:dynamics_cp} shows that the model pre-trained on CP- newborn led to faster convergence and smaller changes in the representations compared with the model pre-trained on Hippocampus. We should point out that for all three training trials in this experiment, the test accuracy of the final models were very close. Therefore, the difference is mainly in terms of the convergence speed. Nonetheless, this experiment suggests that the more similar the source domain is to the target domain, the faster will be the convergence of the model on the target task and less significant will be the changes in the representations during fine-tuning.

\begin{figure*}[!htb]
  \centering
  \centerline{\includegraphics[width=\textwidth]{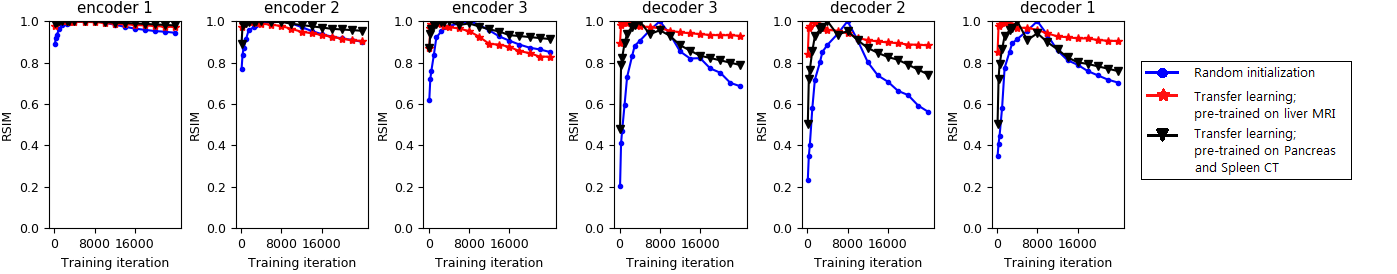}}
\caption{Evolution of learned representations for segmentation of the Liver-CT dataset. The plots show RSIM values computed between the representations at each time point with the convergence epoch. Convergence epoch can be identified as the point where RSIM=1.}
\label{fig:dynamics_liverct}
\end{figure*}

We show another example of the evolution of representations for segmentation of Liver-CT dataset in Figure \ref{fig:dynamics_liverct}. This figure displays some of the observations explained above for cortical plate segmentation. In one of the transfer learning trials in this experiment, we pre-trained the model on Pancreas CT and Spleen CT datasets, which share the imaging modality (CT) with the target task. In the other transfer learning experiment, we pre-trained on the three liver MRI datasets (see Table \ref{table:data}). An interesting observation is that, compared with the model pre-trained on Pancreas and Spleen CT datasets, the model pre-trained on liver MRI went through less changes in the decoder section during fine-tuning. This makes intuitive sense because the network pre-trained on liver MRI had to learn many high-level shape representations that were relevant for liver CT segmentation as well. On the other hand, in terms of the encoder representations, networks pre-trained on MRI and CT images were not very different. This may seem counter-intuitive because one may expect that for segmentation of liver in CT a model pre-trained on CT images should go through less changes in the encoder section during fine-tuning. We will further explain this observation later in this section.

\begin{figure}[!htb]
  \centering
  \centerline{\includegraphics[width=0.5\textwidth]{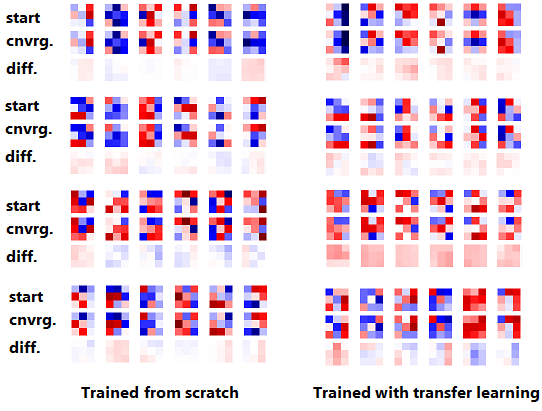}}
\caption{Selected filters from different sections of the network at the start of training and convergence, and their differences. In this experiment, the network was trained on the TSC dataset and for transfer learning the model was pre-trained on the BRATS dataset. From top to bottom, filters belong to encoder-1, encoder-3, decoder-1, and decoder-3 layers (see Figure \ref{fig:network}).}
\label{fig:filters_brats}
\end{figure}

Another interesting observation from both Figures \ref{fig:dynamics_cp} and \ref{fig:dynamics_liverct} is that, both with transfer learning and with training from scratch, the encoder representations changed much less than the decoder representations. The early encoder layer representations, in particular, changed very little even when trained from random initializations. To further confirm this observation, in Figure \ref{fig:filters_brats} we show selected convolutional filters from the encoder and decoder sections of the network at the beginning and end of training, both for training from scratch and for transfer learning. The most striking observation is that the filters change very little during training, and the shape of the filters remain almost unchanged. This example is for brain lesion segmentation on the TSC dataset, and the transfer learning was performed on the BRATS dataset. When training from scratch, the network weights at convergence look random and still very close to the weights at initialization. Some of the filters of the network pre-trained on BRATS look more ``organized" as edge detectors, but still there are random-looking filters, and during fine-tuning on the TSC dataset weights changed very little. We made similar observations in experiments with other datasets. In our experiments, the average relative change in the norm of the filters at convergence was typically below 25\% of the filter norm at the start of training. We also found that, in most cases, the filters in the intermediate layers of the network changed more than the early and late layer filters. This may be related to the observation in previous studies that optimization of the middle layers is more difficult \cite{yosinski2014}. The reason why decoder representations change much more than encoder representations (as shown in Figures \ref{fig:dynamics_cp} and \ref{fig:dynamics_liverct}) is because the change in each representation is the ``cumulative" effect of the changes in all its preceding convolutional layers.

\begin{figure}[!htb]
  \centering
  \centerline{\includegraphics[width=0.5\textwidth]{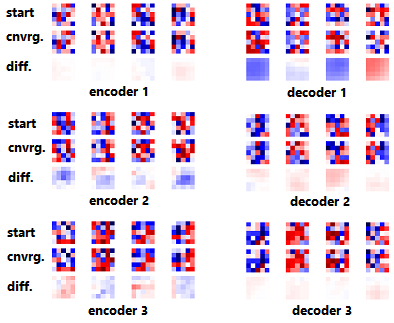}}
\caption{Selected filters from encoder and decoder sections of V-Net trained on Liver-CT dataset at the start of training, convergence, and their difference.}
\label{fig:filters_vnet}
\end{figure}

To experiment with a different network architecture, we implemented the V-Net \cite{milletari2016}. Compared with our network (Figure \ref{fig:network}), V-Net has convolutional filters of size 5 and fewer connections. In terms of segmentation accuracy, V-Net consistently under-performed compared with our network. On some data with fine patterns such as cortical plate, it performed poorly. Here we consider the Liver-CT dataset, on which V-Net performed well, achieving a mean DSC of 0.95. Figure \ref{fig:filters_vnet} shows selected filters from encoder and decoder sections of V-Net at random initialization and at convergence. Similar to Figure \ref{fig:filters_brats}, filters at convergence have changed little compared with those at initialization and still look random. We further investigate this observation in the following subsection.

\subsection{Segmentation networks with \emph{random} encoders}

The observation that convolutional filters of fully-trained models look random and only slightly change from their initial random values may seem surprising at first. One may speculate that this is due to limited training data. However, we conducted a multi-task segmentation experiment in which we trained a single model to segment 10 of the datasets listed in Table \ref{table:data}, consisting of more than 1200 images. We observed similar patterns as those shown in Figure \ref{fig:filters_brats}. We should point out that, even on small datasets, if we continue training the network well beyond the convergence, the weights will continue to change and become more different from the weights at initialization. However, we are only interested in the ``necessary" change that occurs from the start of training until convergence.

\begin{figure}[!htb]
  \centering
  \centerline{\includegraphics[width=0.5\textwidth]{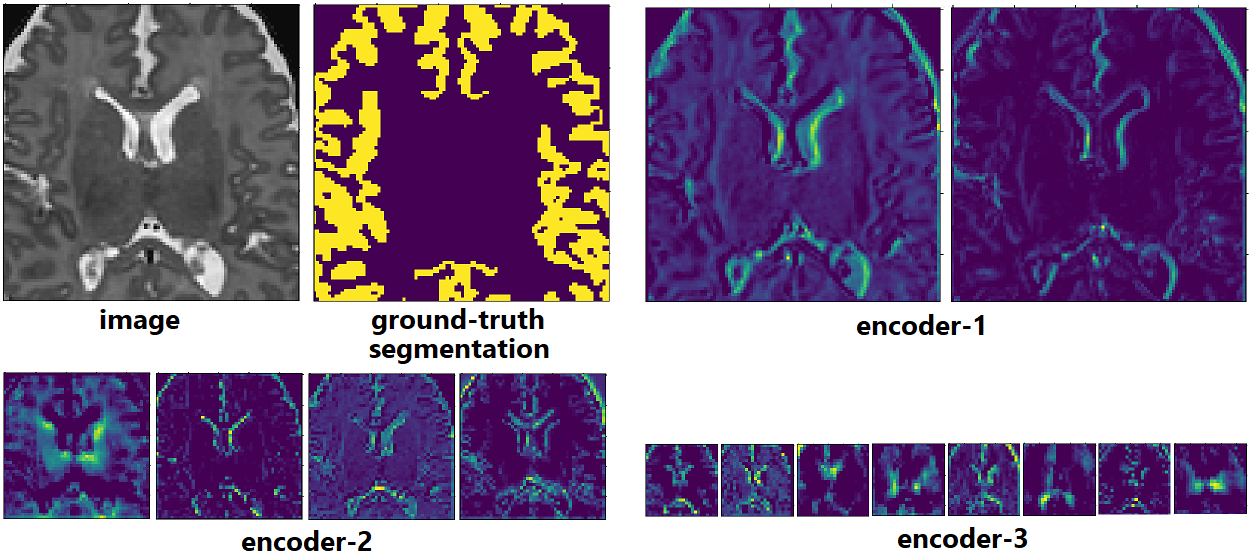}}
\caption{Example feature maps computed with random-looking convolutional filters of a model trained on CP-newborn dataset on an example image patch.}
\label{fig:feature_maps_rand_filter}
\end{figure}

The above observation may seem to contradict the expectation that ``useful" filters such as edge detectors and Gabor-type filters should emerge in the encoder section of the network. However, this observation can be explained by prior studies on neural networks with random weights \cite{saxe2011,cao2018review}. Well before the recent surge of deep learning, studies had shown that neural networks with completely random weights could perform well on various vision tasks \cite{jarrett2009,pinto2009high}. Saxe et al. explained these observations by showing a remarkable response similarity between sinusoidal and random convolutional filters \cite{saxe2011}. Specifically, they showed that for both sinusoidal and random filters, the maximum-response input was in the form of a sinusoid with a frequency equal to the maximum frequency of the filter. We can visually confirm this by looking at the feature maps of the encoder section of our network at convergence. Figure \ref{fig:feature_maps_rand_filter} shows example feature maps of a network trained on CP-newborn dataset. Although the filters of this network looked random (similar to those shown in Figure \ref{fig:filters_brats}), the extracted features do not look random; rather, they embody meaningful low-level and high-level features.

Given the above observations, two questions are worth further investigation. First, how would an FCN with completely random filters (i.e., not undergone any training) in the encoder section perform on medical image segmentation tasks? Second, if a network with random filters is a viable model, can we say anything about the space of the models that can successfully perform a segmentation task, and does transfer learning constrain this space? Below, we report experiments that aim at shedding some light on these questions.

\begin{figure}[!htb]
  \centering
  \centerline{\includegraphics[width=0.5\textwidth]{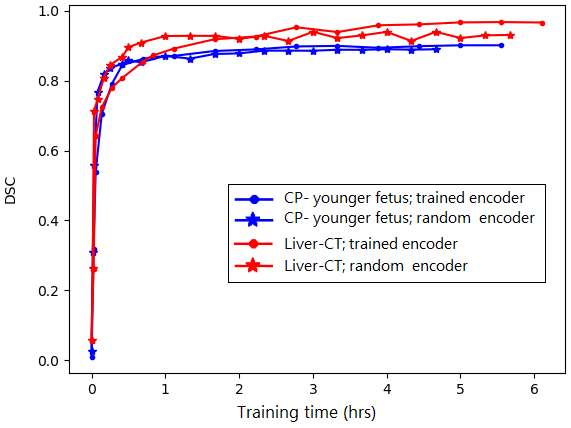}}
\caption{Comparison of a training strategy whereby the encoder section of the network is frozen at its initial random values, only training the decoder section, with the standard strategy of training the entire network.}
\label{fig:frozen_encoder}
\end{figure}

In a set of experiments, we initialized our network at random, and then froze the encoder section of the model, only training the decoder section. Our network (Figure \ref{fig:network}) includes a total of 33 convolutional layers in the encoder section and 18 convolutional layers in the decoder section. Figure \ref{fig:frozen_encoder} shows a comparison of this training strategy with the standard approach of training the entire network on two datasets, i.e., CP-younger fetus and Liver-CT. The time to run one optimization operation on our GPU was 1.08 and 0.63 seconds, respectively, for optimizing the entire network and optimizing the decoder alone. Therefore, the horizontal axis is shown in hours, rather than iteration count. For Liver-CT dataset, the test DSC at convergence was 0.967 and 0.940, respectively, for the experiments with the trained encoder and with the random frozen encoder. For CP-younger fetus dataset, the DSC at convergence was 0.896 and 0.884, respectively, for experiment with trained encoder and with random frozen encoder. This indicates a small drop in performance when the encoder section was frozen at its initial random state. On the other hand, the network with frozen encoder converged in shorter time compared to the network with trained encoder.

The above experiment suggests that \emph{the encoder section does not have to be trained}. We cannot claim that this would be the case for every medical image segmentation task. Nonetheless, we made the same observation on many experiments with various datasets in Table \ref{table:data}. As a concrete example with a very different dataset than the two datasets used above, in an experiment with the TSC dataset from one of the five centers, we obtained DSC and F1 score, respectively, of 0.678 and 0.758 when the entire network was trained and 0.670 and 0.758 when the encoder section was frozen at its random initialization. These observations clearly challenge the common belief that the encoder filters have to learn data/task-specific features.

Based on our observations with FCNs with random encoders and the response of random convolutional filters discussed above, one can describe a possible operation mode of these networks as follows. The encoder filters extract a set of useful representations from the image. Although the filters might be random-looking, the representations embody relevant features such as edges in early layers and high-level features in deeper layers. Given that filters are initialized independently at random, these feature maps will constitute a diverse and rich set of representations. The decoder section learns to compute the segmentation label based on these representations.

We conducted other experiments to assess the similarity of FCNs trained from scratch compared with similarity of FCNs trained via transfer learning. Table \ref{table:similarities} shows the results of such an experiment with CP-younger fetus dataset. For this experiment, we trained 1) 10 networks with different random initializations, and 2) 10 networks trained with transfer learning, each initialized from a different model pre-trained on CP-newborn. We then computed the similarity between pairs of networks trained from scratch ($\text{RSIM}(R.I., R.I.)$), pairs of networks trained with transfer learning ($\text{RSIM}(T.L., T.L.)$), as well as similarity between pairs of networks trained using the two different strategies ($\text{RSIM}(R.I., T.L.)$). Table \ref{table:similarities} shows these similarities for the six layers shown in Figure \ref{fig:network}.

\begin{table*}[!htb]
\scriptsize
 \caption{\small{Similarity of representations between different layers of pairs of networks trained from random initialization ($R.I.$) and via transfer learning ($T.L.$).}}
    \label{table:similarities}
\begin{tabular}{ L{2.5cm}  C{2.1cm} C{2.1cm} C{2.1cm} C{2.1cm} C{2.1cm} C{2.1cm} }
\thickhline
 &  encoder-1 & encoder-2 & encoder-3 &  decoder-1 & decoder-2 & decoder-3   \\ \thickhline
$\text{RSIM}(R.I., R.I.)$  & $ 0.247 \pm 0.045 $ & $ 0.358 \pm 0.034 $ & $ 0.405 \pm 0.018 $ & $ 0.463 \pm 0.046 $ & $ 0.540 \pm 0.064 $ & $ 0.398 \pm 0.041 $   \\
$\text{RSIM}(T.L., T.L.)$  & $ 0.243 \pm 0.010 $ & $ 0.370 \pm 0.009 $ & $ 0.433 \pm 0.005 $ & $ 0.510 \pm 0.018 $ & $ 0.492 \pm 0.047 $ & $ 0.333 \pm 0.050 $   \\
$\text{RSIM}(R.I., T.L.)$  & $ 0.287 \pm 0.019 $ & $ 0.386 \pm 0.014 $ & $ 0.460 \pm 0.021 $ & $ 0.597 \pm 0.024 $ & $ 0.636 \pm 0.046 $ & $ 0.534 \pm 0.057 $   \\
\thickhline
\end{tabular}
\end{table*}

Table \ref{table:similarities} shows that models trained via transfer learning are as different from each other as models learned from scratch. Statistical t-tests comparing $\text{RSIM}(R.I., R.I.)$ with $\text{RSIM}(T.L., T.L.)$ showed that only encoder-3 and decoder-1 layers (i.e., the most intermediate of the six layers) were different at $p=0.05$. On the other hand, $\text{RSIM}(R.I., R.I.)$ and $\text{RSIM}(R.I., T.L.)$ were different at $p=0.05$ on all six layers. Similarly, $\text{RSIM}(T.L., T.L.)$ were different from $\text{RSIM}(R.I., T.L.)$ on all six layers at $p=0.05$. These results indicate that, at convergence, the diversity of networks trained using transfer learning is as high as that of networks trained from scratch. However, it also shows that networks trained from scratch form a different ``cluster" than networks trained via transfer learning. This is because $\text{RSIM}(R.I., R.I.)$ and $\text{RSIM}(T.L., T.L.)$ are smaller than $\text{RSIM}(R.I., T.L.)$.

Finally, we quantified feature reuse using the method proposed in \cite{raghu2019transfusion}. We trained 10 models with random initialization and 10 with transfer learning and computed the similarity between the network layers at the start of training and at convergence. We then computed the difference between the average of these similarities for the models trained from scratch and models trained with transfer learning. As proposed in \cite{raghu2019transfusion}, we use this difference as a measure of feature reuse in each layer. Table \ref{table:feature_reuse} shows the computed feature reuse for four such experiments.

\begin{table*}[!htb]
\scriptsize
 \caption{\small{Feature reuse in different network layers in four transfer learning experiments.}}
    \label{table:feature_reuse}
\begin{tabular}{ L{6.5cm}  C{1.4cm} C{1.4cm} C{1.4cm} C{1.4cm} C{1.4cm} C{1.4cm}  }
\thickhline
 &  encoder-1 & encoder-2 & encoder-3 &  decoder-1 & decoder-2 & decoder-3   \\ \thickhline
Liver-CT, transfer learning from liver MRI  &  0.123 & 0.210 & 0.339 & 0.690 & 0.611 & 0.530  \\
Liver-CT, transfer learning from Pancreas and Spleen CT  &  0.156 & 0.162 & 0.242 & 0.214 & 0.203 & 0.107  \\ \hline
CP- younger fetus; transfer learning from CP- newborn & 0.164 & 0.267 & 0.394 & 0.392 & 0.484 & 0.611 \\
CP- younger fetus; transfer learning from Hippocampus & 0.120 & 0.212 & 0.305 & 0.239 & 0.203 & 0.224 \\
\thickhline
\end{tabular}
\end{table*}

Our observations are quite different from those reported for 2D medical image classification with transfer learning from ImageNet \cite{raghu2019transfusion}. In particular, \cite{raghu2019transfusion} found that feature reuse was largest (at less than 0.20) in the first layers and that it decreased monotonically from bottom layers to top layers. We observe an exact opposite trend in some of our experiments; feature reuse can increase from early encoder layers to deeper decoder layers. Moreover, we observe much higher feature reuse of up to 0.690. These observations can guide us in devising transfer learning strategies. For example, Table \ref{table:feature_reuse} shows high feature reuse in the decoder section when a model trained on liver MRI was fine-tuned for Liver-CT segmentation. This indicates that fine-tuning of decoder layers may be unnecessary. Indeed we found that a mean test DSC of above 0.95 on Liver-CT dataset could be achieved by fine-tuning only the last two layers of the model pre-trained on liver MRI. Equally accurate models could be trained by keeping the decoder and output layers fixed and only training the encoder section of the network. Similarly, for fine-tuning the model pre-trained on CP-newborn for segmentation of CP-younger fetus, we could achieve a competitive DSC of 0.89 by fine-tuning only the last two layers of the decoder section or only the encoder section.

\section{Conclusion}

Our experimental results show that for segmenting organs such as liver or brain cortical plate, transfer learning has a small effect on the segmentation accuracy. One may argue that the segmentation accuracy achieved by FCNs is close to optimal and the remaining gap in performance (DSC gap of 0.03 for liver and 0.10 for cortical plate) may be, in part, due to error in the training/test labels. Our experiments with pancreas and brain lesion segmentation seem to give some credence to this hypothesis. In those experiments, the segmentation accuracy (e.g., in terms of DSC) was much lower and we observed larger gains with transfer learning. We observed large gains in segmentation accuracy when images in the target domain were of a different/lower quality and small in number (Table \ref{table:tl_center5_table}). This is important since medical image datasets of varying quality are common in clinical and research applications. We also observed that transfer learning and learning with a frozen encoder reduced the convergence time. This could be useful when training time is critical or in hyperparameter/architecture search, where one would like to compare a large number of models or hyperparameter settings.

We showed that random filters extracted a rich set of useful features, and that quite accurate models could be built by training only the decoder section of the model. We further demonstrated that, depending on the source and target domains, feature reuse in transfer learning can be more significant in deeper layers and in the decoder section of the network. We showed examples of how these observations could be used in devising transfer learning strategies.

There are important research questions that we did not pursue in this paper because of space limitation. For example, we only used a scalar measure of similarity to compare networks trained from scratch with networks trained via transfer learning. One may gain more insight by comparing these networks in terms of the projection directions mentioned in Section \ref{cca_section}. Future works can also assess transfer learning for other medical image analysis tasks, such as detection and classification. Given the inherent differences between those tasks and voxel-wise segmentation, separate studies investigating the role of transfer learning in those tasks are well justified.

\section*{Acknowledgment}

The DHCP dataset is provided by the developing Human Connectome Project, KCL-Imperial-Oxford Consortium funded by the European Research Council under the European Union Seventh Framework Programme (FP/2007-2013) / ERC Grant Agreement no. [319456]. We are grateful to the families who generously supported this trial.

\bibliographystyle{IEEEtran}
\bibliography{davoodreferences}

\end{document}